\pdfoutput=1
\documentclass{article} 
\usepackage{nips13submit_e,times}
\usepackage{graphicx}
\usepackage{subfig}
\usepackage{hyperref}
\usepackage{amsmath}
\usepackage{algorithm}
\usepackage{multirow}
\usepackage[noend]{algpseudocode}
\usepackage{multirow,multicol, makecell, booktabs}
\usepackage{url}

\makeatletter
\def\BState{\State\hskip-\ALG@thistlm}
\makeatother

\title{Word2Bits - Quantized Word Vectors}

\author{
Maximilian Lam\\
\texttt{maxlam@stanford.edu} \\
}

\nipsfinalcopy 

\begin{document}

\maketitle

\begin{abstract}
Word vectors require significant amounts of memory and storage, posing
issues to resource limited devices like mobile phones and GPUs. We
show that high quality quantized word vectors using 1-2 bits per
parameter can be learned by introducing a quantization function into
Word2Vec. We furthermore show that training with the quantization
function acts as a regularizer. We train word vectors on English
Wikipedia (2017) and evaluate them on standard word similarity and
analogy tasks and on question answering (SQuAD). Our quantized word
vectors not only take 8-16x less space than full precision (32 bit)
word vectors but also outperform them on word similarity tasks and
question answering.
\end{abstract}

\section{Introduction}
Word vectors are extensively used in deep learning models for natural
language processing. Each word vector is typically represented as a
300-500 dimensional vector, with each parameter being 32 bits. As
there are millions of words, word vectors may take up to 3-6 GB of
memory/storage -- a massive amount relative to other portions of a
deep learning model[25]. These requirements pose issues to
memory/storage limited devices like mobile phones and GPUs.

Furthermore, word vectors are often re-trained on application specific
data for better performance in application specific domains[27]. This
motivates directly learning high quality compact word representations
rather than adding an extra layer of compression on top of pretrained
word vectors which may be computational expensive and degrade
accuracy.

Recent trends indicate that deep learning models can reach a high
accuracy even while training in the presence of significant noise and
perturbation[5, 6, 9, 28, 32]. It has furthermore been shown that high quality
quantized deep learning models for image classification can be learned
at the expense of more training epochs[9]. Inspired by these trends we
ask: can we learn high quality word vectors such that each parameter
is only one of two values, or one of four values (quantizing to 1 and
2 bits respectively)?

To that end we propose learning quantized word vectors by introducing
a quantization function into the Word2Vec loss formulation -- we call
our simple change Word2Bits. While introducing a quantization function
into a loss function is not new, to the best of our knowledge it is
the first time it has been applied to learning compact word representations.

In this report we show that
\begin{enumerate}
\item[$\bullet$]

  It is possible to train high quality quantized word vectors which
  take 8x-16x less storage/memory than full precision word
  vectors. Experiments on both intrinsic and extrinsic tasks show that
  our learned word vectors perform comparably or even better on many tasks.

\item[$\bullet$]

  Standard Word2Vec may be prone to overfitting; the quantization
  function acts as a regularizer against it.

\end{enumerate}

\section{Related Work}
Word vectors are continuous representations of words and are used by
most deep learning NLP models. Word2Vec, introduced by Mikolov's
groundbreaking papers[7, 8], is an unsupervised neural network algorithm for
learning word vectors from textual data. Since then, other
groundbreaking algorithms (Glove, FastText) [2, 11] have been proposed to
learn word vectors using other properties of textual data. As of 2018
the most widely used word vectors are Glove, Word2Vec and
FastText. This work focuses on how to learn memory/storage efficient
word vectors through quantized training -- specifically our approach
extends Word2Vec to output high quality quantized word vectors.

Learning compact word vectors is related to learning compressed neural
networks. Finding compact representations of neural
networks date back to the 90's and include techniques like network
pruning[24, 25], knowledge distillation[29], deep compression[25] and
quantization[9]. More recently, algorithmic and hardware advances have
allowed training deep models using low precision floating-point and
arithmetic operations[22, 26] -- this is also referred to as
quantization. To distinguish between quantized training with low
precision arithmetic/floats from quantized training with full
precision arithmetic/floats but constrained values we term the first
physical quantization and the latter virtual quantization.

Our technical approach follows that of neural network quantization for
image classification[9], which does virtual quantization by
introducing a sign function (a 1 bit quantization function) into the
training loss function. The actual technique of backpropagating through a
discrete function (the quantization function) has been thoroughly
explored by Hinton[10] and Bengio[30].

Application wise, various techniques exist to compress word
embeddings. These approaches involve taking pre-trained word
vectors and compressing them using dimensionality reduction,
pruning[25], or more complicated approaches like deep compositional
coding[25]. Such techniques add an extra layer of computation to
compress pre-trained embeddings and may degrade word vector
performance[25].

To the best of our knowledge, current traditional methods of obtaining
compact word vectors involve adding an extra layer of computation to
compress pretrained word vectors[1, 23, 25] (as described previously). This
may incur computational costs which may be expensive in context of
retraining word vectors for application specific purposes and may
degrade word vector performance[25]. Our proposed approach of directly
learning quantized word vectors from textual data may amend these
issues and is an alternative method of obtaining compact high quality
word vectors. Note that these traditional compression methods may
still be applied on the learned quantized word vectors.

\section{Word2Bits - Quantized Word Embeddings}
\subsection{Background}
Our approach utilizes the Word2Vec formulation of learning word vectors. There
are two Word2Vec algorithms: Skip Gram Negative Sampling (SGNS) and
Continuous Bag of Words (CBOW)[7] -- our virtual quantization
technique utilizes CBOW with negative sampling. The CBOW negative
sampling loss function minimizes

$$
J(u_o, \hat{v}_c) = -log(\sigma(u_o^T\hat{v}_c)) - \sum_{i=1}^{k} log(\sigma(-u_i^T\hat{v}_c))
$$

where

$$
u_o = \mbox{ vector of center word with corpus position } o
$$
$$
\hat{v}_c = \frac{1}{2w}\sum_{-w+o \leq i \leq w+o,i \neq o} v_i \mbox{  where } v_i \mbox{ is vector for context word, } w \mbox{ is window size, a hyperparameter}
$$
$$
k = \mbox{ number of negative samples, a hyperparameter}
$$

Intuitively, minimizing this loss function optimizes vectors of words
that occur in similar contexts to be ``closer'' to each other, and
pushes vectors whose contexts are different ``away''. Specifically
CBOW with negative sampling tries to predict the center word from
context words.

Technically, to optimize this loss function, for each window of words:

\begin{enumerate}

\item[$\bullet$] Identify the center word's vector $u_o$ within the window
\item[$\bullet$] Compute the average of the context words $\hat{v}_c = \frac{1}{2w} \sum_{-w+o \leq i \leq w+o, i \neq o} v_i$ given window size $w$
\item[$\bullet$] Draw $k$ negative samples $u_1, u_2, .., u_k$ according to a sampling distribution [1].
\item[$\bullet$] Compute loss $J(u_o, \hat{v}_c) = -log(\sigma(u_o^T\hat{v}_c)) - \sum_{i=1}^{k} log(\sigma(-u_i^T\hat{v}_c))$
\item[$\bullet$] Update center word vector $u_o$ with gradient $\frac{\partial J(u_o, \hat{v}_c)}{\partial u_o}$
\item[$\bullet$] Update negative word vector $u_i$ with gradient $\frac{\partial J(u_o, \hat{v}_c)}{\partial u_i}$
\item[$\bullet$] Update context word vector $v_i$ with gradient $\frac{\partial J(u_o, \hat{v}_c)}{\partial v_i}$
\end{enumerate}

Center vectors $u_i$ and context vectors $v_j$ are stored full
precision. The final word vectors are the sums of the context and
center vectors $u_i + v_i$ for each corresponding word. The resulting
vectors are full precision.

\subsection{Word2Bits Approach}
To learn quantized word vectors we introduce virtual quantization into the CBOW loss function:
$$
J_{quantized}(u^{(q)}_o, \hat{v}^{(q)}_c) = -log(\sigma((u^{(q)}_{o})^{T} \hat{v}^{(q)}_c)) - \sum_{i=1}^{k} log(\sigma((-u^{(q)}_i)^T\hat{v}^{(q)}_c))
$$

where

$$
u^{(q)}_o = Q_{bitlevel}(u_o)
$$

$$
\hat{v}^{(q)}_c = \sum_{-w+o \leq i \leq w+o,i \neq o} Q_{bitlevel}(v_i)
$$

$$
Q_{bitlevel}(x) = \mbox{ quantization function to quantize to } bitlevel \mbox{ bits}
$$

The following quantization functions were used (chosen based on what worked best)

\[
Q_1(x) =
\begin{cases}
  \frac{1}{3} & x \geq 0\\
  -\frac{1}{3} & x < 0
\end{cases}
\]

\[
Q_2(x) =
\begin{cases}
  \frac{3}{4} & x > \frac{1}{2}\\
  \frac{1}{4} & 0 \leq x \leq \frac{1}{2}\\
  -\frac{1}{4} & -\frac{1}{2} \leq x < 0\\
  -\frac{3}{4} & x < -\frac{1}{2}
\end{cases}
\]

Since $Q_{bitlevel}$ is a discrete function, its derivative is
undefined at some points and 0 at others. To solve this we simply set
the derivative of $Q_{bitlevel}$ to be the identity function:

$$
\frac{\partial Q_{bitlevel}(x)}{\partial x} = I
$$

This is also known as Hinton's straight-through estimator[10].

The final gradient updates reduce to Word2Vec updates. They are:
$$
\mbox{For center word } u_o \textbf{: } \frac{\partial J_{quantized} (u^{(q)}_o, \hat{v}^{(q)}_c)}{\partial u_o} = \frac{\partial J_{quantized} (u^{(q)}_o, \hat{v}^{(q)}_c)}{\partial u^{(q)}_o}
$$
$$
\mbox{For negative word } u_i \textbf{: } \frac{\partial J_{quantized} (u^{(q)}_o, \hat{v}^{(q)}_c)}{\partial u_i} = \frac{\partial J_{quantized} (u^{(q)}_o, \hat{v}^{(q)}_c)}{\partial u^{(q)}_i}
$$
$$
\mbox{For context word } v_i \textbf{: } \frac{\partial J_{quantized} (u^{(q)}_o, \hat{v}^{(q)}_c)}{\partial v_i} = \frac{\partial J_{quantized} (u^{(q)}_o, \hat{v}^{(q)}_c)}{\partial v^{(q)}_i}
$$

Like in the standard algorithm, we optimize $J_{quantized}$ with
respect to $u_i$ and $v_j$ over a corpus of text. The final vector for
each word is $Q_{bitlevel}(u_i + v_i)$; thus each parameter is one of
$2^{bitlevel}$ values and takes $bitlevel$ bits to represent.

Intuitively, although we are still updating $u_i$ and $v_j$ (full
precision vectors), we are now optimizing their quantized
counterparts $Q_{bitlevel}(u_i)$ and $Q_{bitlevel}(v_j)$ to capture
the same corpus statistics as regular word vectors. While we are still
training with full precision 32-bit arithmetic operations and 32-bit floating
point values, the final word vectors we save to disk are
quantized.

\section{Experiments and Results}
\subsection{Intrinsic Experiments - Word Similarity and Analogy}

\textbf{Word Vector Training Methodology}

We train word vectors with varying levels of precision and dimension
on the 2017 English Wikipedia dump (24G of text). We normalize the
text similar to FastText[2], however we keep the text case
sensitive. We train all word vectors for 25 epochs. We use the
following hyperparameters: window size = 10, negative sample size =
12, min count = 5, subsampling = 1e-4, learning rate = .05 (which is
linearly decayed to 0.0001). Our final vocabulary size is 3.7 million
after filtering words that appear less than min count = 5 times. In
our intrinsic experiments we additionally report the scores of
thresholded vectors (denoted T1) which are computed by taking
trained full precision vectors and applying the 1-bit quantization function
on them.

\textbf{Test Datasets and Evaluation}

Our evaluation procedure follows that of [4]. We use six datasets to
evaluate word similarity and two datasets to evaluate word
analogy. The word similarity test datasets are: WordSim353 Similarity
[16], WordSim353 Relatedness [17], MEN [18], Mechanical Turk [19], Rare
Words [20] and Simlex[21]. The word analogy test datasets are Google's
analogy dataset [9] and MSR's analogy dataset[9]. We modify Google's
analogy dataset by uppercasing the first character of proper nouns (as
we are training case sensitive word vectors). To evaluate word
similarity, word vectors are ranked by cosine similarity; the reported
score is correlation with human rankings[4]. To answer word analogy
questions we use two methods: 3CosAdd (Add) and 3CosMul (Mul) as
detailed in [4]; the reported score is the percentage of questions for
which the argmax vector is the correct answer.

\textbf{Results}

Table 1 shows results of the full intrinsic evaluation. These data
indicate that quantized word vectors perform comparably with full
precision word vectors on many intrinsic tasks. Interestingly,
quantized word vectors outperform full precision vectors on word
similarity tasks, but do worse on word analogy tasks. Thresholded word
vectors perform consistently worse than their full precision
counterparts across all tasks.

\begin{table}[t]
\caption{Word similarity and analogy results}
\label{intrinsic-table}
\begin{center}
\resizebox{\textwidth}{!}{\begin{tabular}{ |c|cc|cccccc|cc| }
\hline
\makecell{Word Vector Type} &
\makecell{Bits per \\ parameter} &
\makecell{Dimension} &
\makecell{WordSim \\ Similarity} &
\makecell{WordSim \\ Relatedness} &
\makecell{MEN} &
\makecell{M. Turk} &
\makecell{Rare \\ Words} &
\makecell{SimLex} &
\makecell{Google \\ Add / Mul} &
\makecell{MSR \\ Add / Mul} \\
\hline

\multirow{4}{6em}{Full Precision}
& 32 & 200 & .740 & .567 & .716 & .635 & .403 & .317 & .706/.702 & .447/.447\\
& 32 & 400 & .735 & .533 & .720 & .623 & .408 & .335 & \textbf{.722}/.734 & \textbf{.473}/.486\\
& 32 & 800 & .726 & .500 & .713 & .615 & .395 & .337 & .719/.735 & .471/\textbf{.489}\\
& 32 & 1000 & .741 & .529 & .745 & .617 & .400 & .358 & .664/.675 & .423/.434\\
\hline
\multirow{4}{6em}{\makecell{Thresholded}}
& T1 & 200 & .692 & .480 & .668 & .575 & .347 & .288 & .371/.369 & .186/.182\\
& T1 & 400 & .677 & .446 & .686 & .581 & .369 & .321 & .533/.540 & .286/.292\\
& T1 & 800 & .728 & .494 & .692 & .576 & .383 & .338 & .599/.609 & .333/.346\\
& T1 & 1000 & .689 & .504 & .694 & .551 & .358 & .342 & .521/.520 & .303/.305\\
\hline
\multirow{6}{6em}{\makecell{Quantized}}
& 1 & 800 & .772 & .653 & .746 & .612 & .417 & .355 & .619/.660 & .395/.390\\
& 1 & 1000 & .768 & \textbf{.677} & .756 & .638 & \textbf{.425} & .372 & .650/.660 & .371/.408\\
& 1 & 1200 & \textbf{.781} & .628 & .765 & \textbf{.643} & .415 & .379 & .659/.692 & .391/.429\\
& 2 & 400 & .752 & .604 & .741 & .616 & .417 & .373 & .666/.690 & .396/.418\\
& 2 & 800 & .776 & .634 & \textbf{.767} & .642 & .390 & \textbf{.403} & .710/.739 & .418/.460\\
& 2 & 1000 & .752 & .594 & .764 & .602 & .362 & .387 & .720/\textbf{.750} & .436/.482\\
\hline

\hline

\end{tabular}}
\end{center}
\end{table}

\subsection{Extrinsic Experiments - Question Answering}

\textbf{Word Vector Training Methodology}

We use the same word vectors as the intrinsic tasks. Word vectors were
trained on 2017 English Wikipedia (24G of text) on normalized text[2]
keeping case sensitivity. All word vectors were trained for 25 epochs
and with the following hyperparameters: window size = 10, negative
sample size = 12, min count = 5, subsampling = 1e-4, learning rate =
.05 (which is linearly decayed to 0.0001). Our final vocabulary size
is 3.7 million after filtering words that appear less than min count =
5 times.

\textbf{SQuAD Model}

Using our word vectors, we train Facebook's official DrQA[14] model
for the Stanford Question Answering task (SQuAD)[13]. Implementation
details and hyperparameters follow[14] with the following differences:
word embeddings are fixed (instead of allowing the top 1000 to be fine
tuned) and the model is trained for 50 epochs (instead of 40). Note
that the DrQA model is trained entirely in full precision.

\textbf{Results}

Table 2 shows the best development F1 scores achieved across training
epochs by full precision vectors and quantized vectors on SQuAD. The
data show that quantized vectors outperform full precision vectors by
around 1 F1 point; the best performing word vector (400 dimensional
2-bit word vectors) uses 100 bytes per word, which is 8x-16x less than
full precision word vectors. Interestingly, there is a sharp drop in
F1 score from 32-bit 800 dimensional vectors (F1=75.31) to 32-bit 1000
dimensional vectors (F1=9.99). Upon inspection of the 32-bit 1000
dimensional word vectors, we found that parameter values had
``exploded'' to large absolute magnitudes
({\raise.17ex\hbox{$\scriptstyle\mathtt{\sim}$}} 1000000). Intrinsic
tasks were unaffected by this phenomena as vectors were normalized
before processing them (unlike the default DrQA code which does not
normalize the word vectors). We believe that normalizing the full
precision 1000 dimensional vectors would yield better scores.

\begin{table}[t]
\caption{DrQA SQuAD results and vector sizes for full precision and quantized word vectors}
\label{intrinsic-table}
\begin{center}
\begin{tabular}{ |c|cc|c|c|}
\hline
\makecell{Word Vector Type} &
\makecell{Bits per \\ parameter} &
\makecell{Dimension} &
\makecell{Bytes per word} &
\makecell{F1} \\
\hline

\multirow{4}{6em}{Full Precision}
& 32 & 200 & 800 & 75.25\\
& 32 & 400 & 1600 & 75.28\\
& 32 & 800 & 3200 & 75.31\\
& 32 & 1000 & 4000 & 9.99\\
\hline
\multirow{6}{6em}{\makecell{Quantized}}
& 1 & 800 & 100 & 76.64\\
& 1 & 1000 & 125 & 76.84\\
& 1 & 1200 & 150 & 76.50\\
& 2 & 400 & 100 & \textbf{77.04}\\
& 2 & 800 & 200 & 76.12\\
& 2 & 1000 & 250 & 75.66\\
\hline

\hline

\end{tabular}
\end{center}
\end{table}

\subsection{Word2Bits and Regularization}

\textbf{Experiment Details}

To understand why quantized word vectors perform consistently better
on word similarity and question answering we train word vectors on
100MB of wikipedia (text8; case insensitive; Matt Mahoney
processed)[31] with the following hyperparameters:

\begin{enumerate}
\item[$\bullet$] Window size = 10
\item[$\bullet$] Negative sample size = 24
\item[$\bullet$] Subsampling = 1e-4
\item[$\bullet$] Min count = 5
\item[$\bullet$] Learning rate = .05 (linearly decayed to .0001)
\item[$\bullet$] Number of training epochs = [1, 10, 25, 50]
\item[$\bullet$] Bits per parameter = [1, 32]
\item[$\bullet$] Dimension = [100, 200, 400, 600, 800, 1000]
\end{enumerate}

For each individual run we track Google analogy score and end training loss.

\textbf{Results and Analysis}

Figure 1a shows training loss and accuracy versus epochs of training
(with vector dimension fixed at 400); figure 1b shows training loss
and accuracy versus vector dimension (with the number of epochs fixed
at 10). Figure 1a indicates that full precision Word2Vec is prone to overfitting with
increased epochs of training; quantized training does not seem to
suffer as much from this. Figure 1b indicates that full precision Word2vec is prone
to overfitting with increased dimensions; quantized training performs
poorly with fewer dimensions and better with larger dimensions. While
100MB is too small a dataset to make a decisive conclusion, the trends
strongly hint that overfitting is an issue for Word2Vec and that
quantized training may be a form of regularization.

\begin{figure}%
    \centering
    \subfloat[Training loss and accuracy vs epochs trained (vector dimension = 400) on 100MB of Wikipedia. Trends show that Word2Vec is prone to overfitting with many epochs of training.]{{\includegraphics[width=6cm]{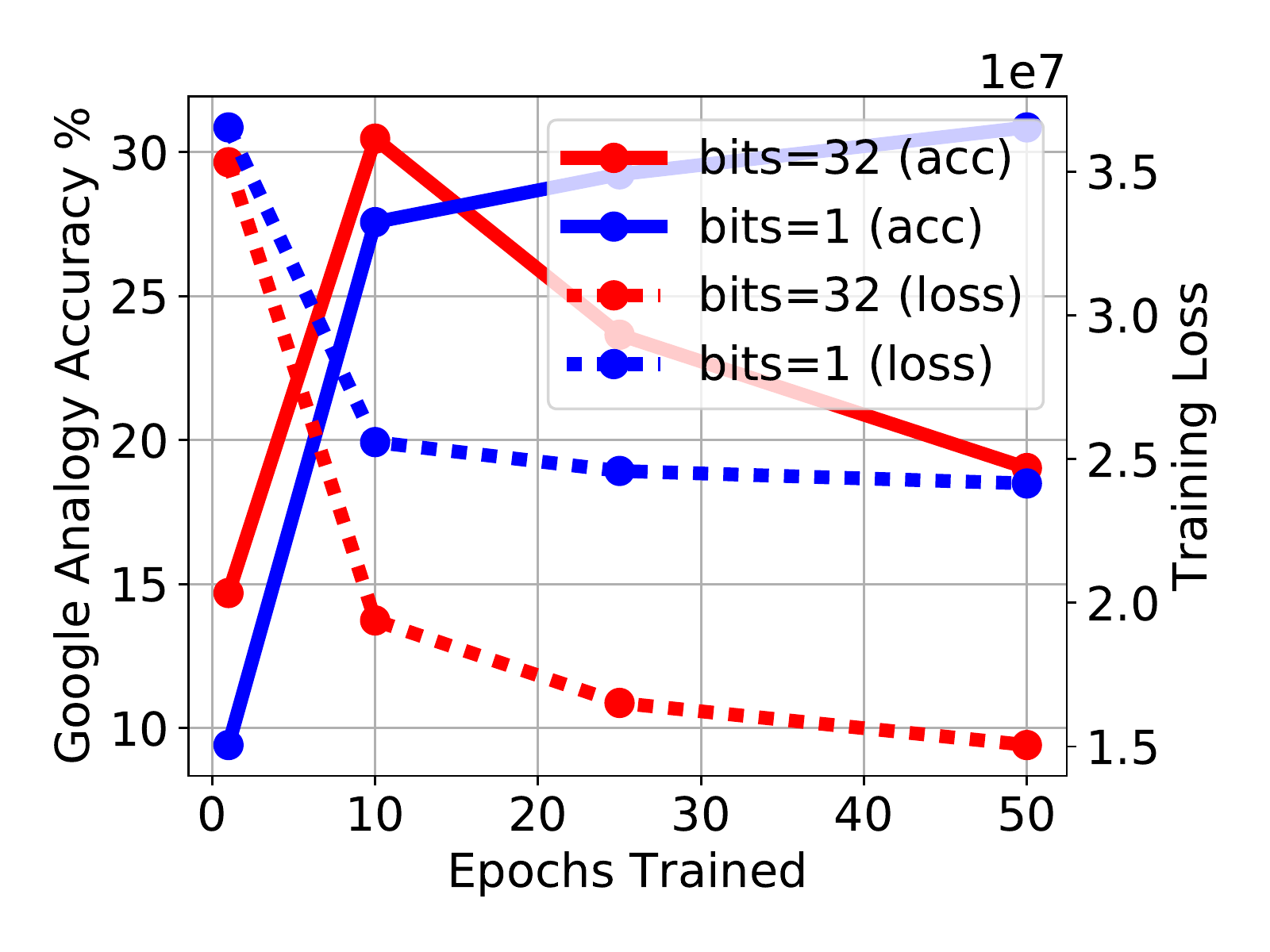}}}%
    \qquad
    \subfloat[Training loss and accuracy vs dimension (epochs trained = 10) on 100MB of Wikipedia. Trends show that overfitting may occur with larger vector dimensions.]{{\includegraphics[width=6cm]{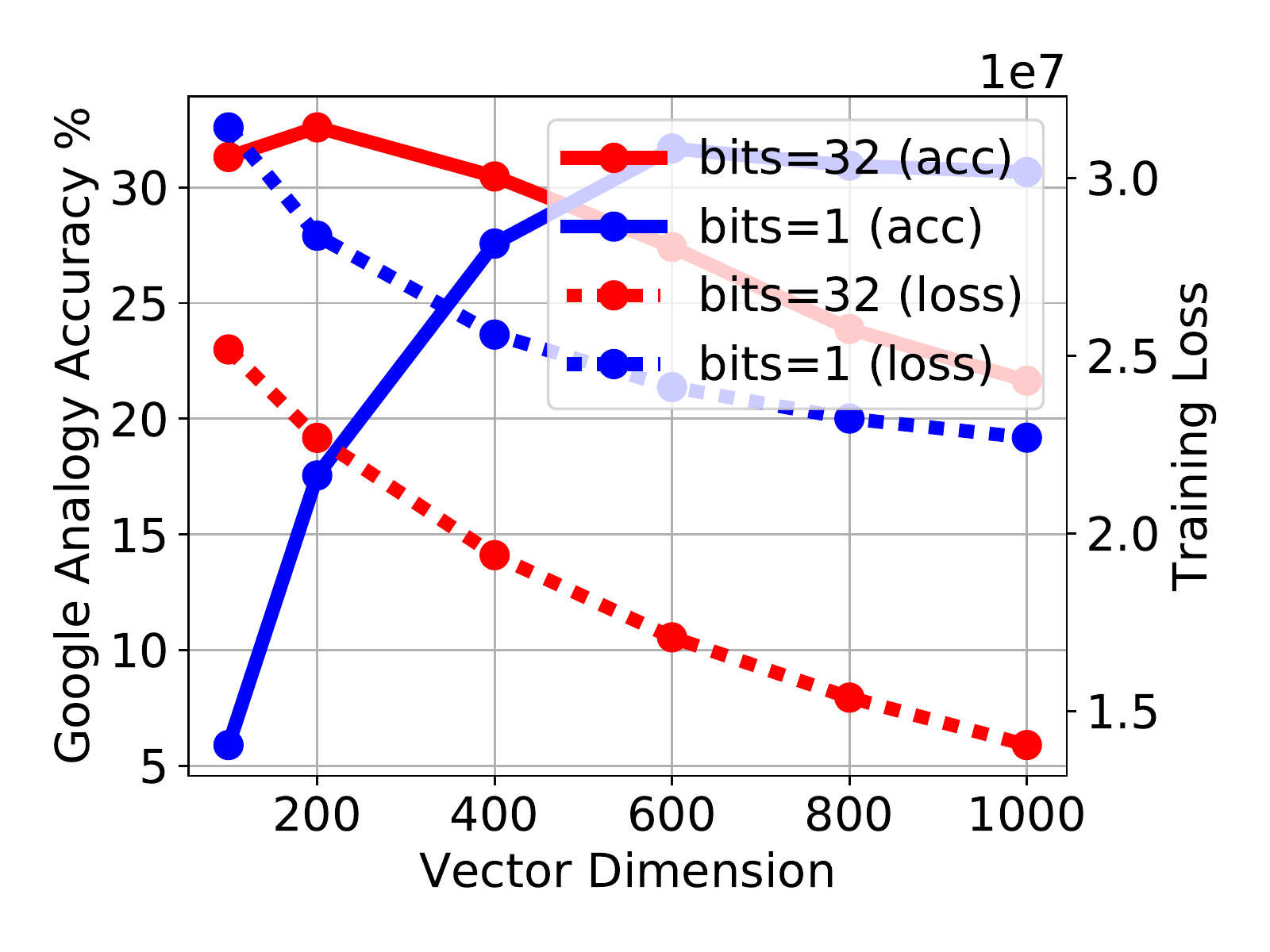} }}%
    \caption{Overfitting in full precision Word2Vec training; regularization in quantized Word2vec training}%
    \label{fig:overfitting}
\end{figure}

\subsection{Word2Bits Visualization}
Figure 2 shows a visualization of 800 dimensional 1 bit word vectors
trained on English Wikipedia (2017). The top 100 closest and furthest
word vectors to the target word vector are plotted. Distance is
measured by dot product; every 5 word vectors are labelled. A
turquoise line separates the 100 closest vectors to the target word
from the 100 furthest vectors (labelled ``...'').  We see that there
are qualitative similarities between word vectors whose words are
related to each other.

\begin{figure}%
    \centering
    \subfloat[]{{\includegraphics[width=6cm]{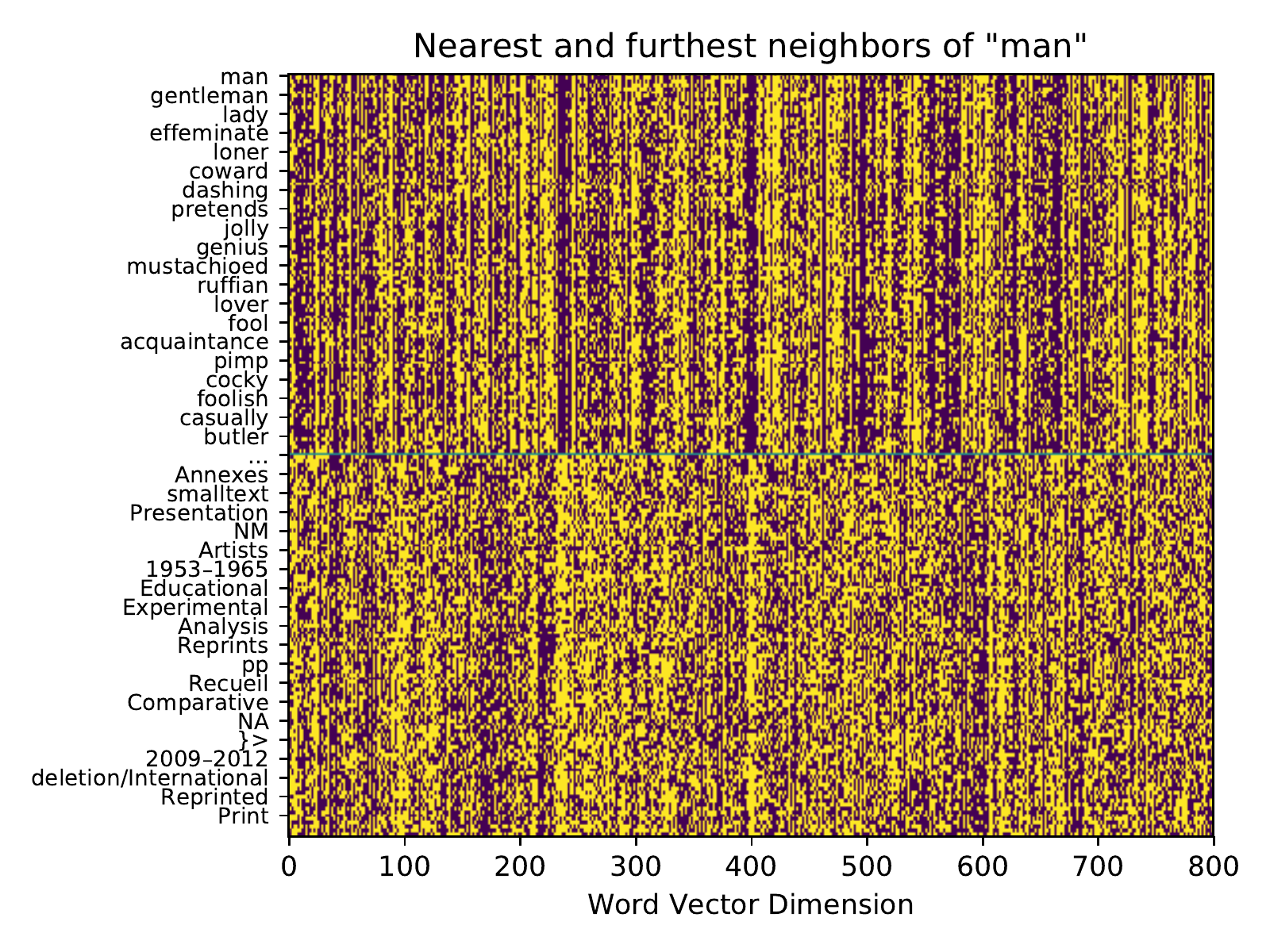}}}%
    \qquad
    \subfloat[]{{\includegraphics[width=6cm]{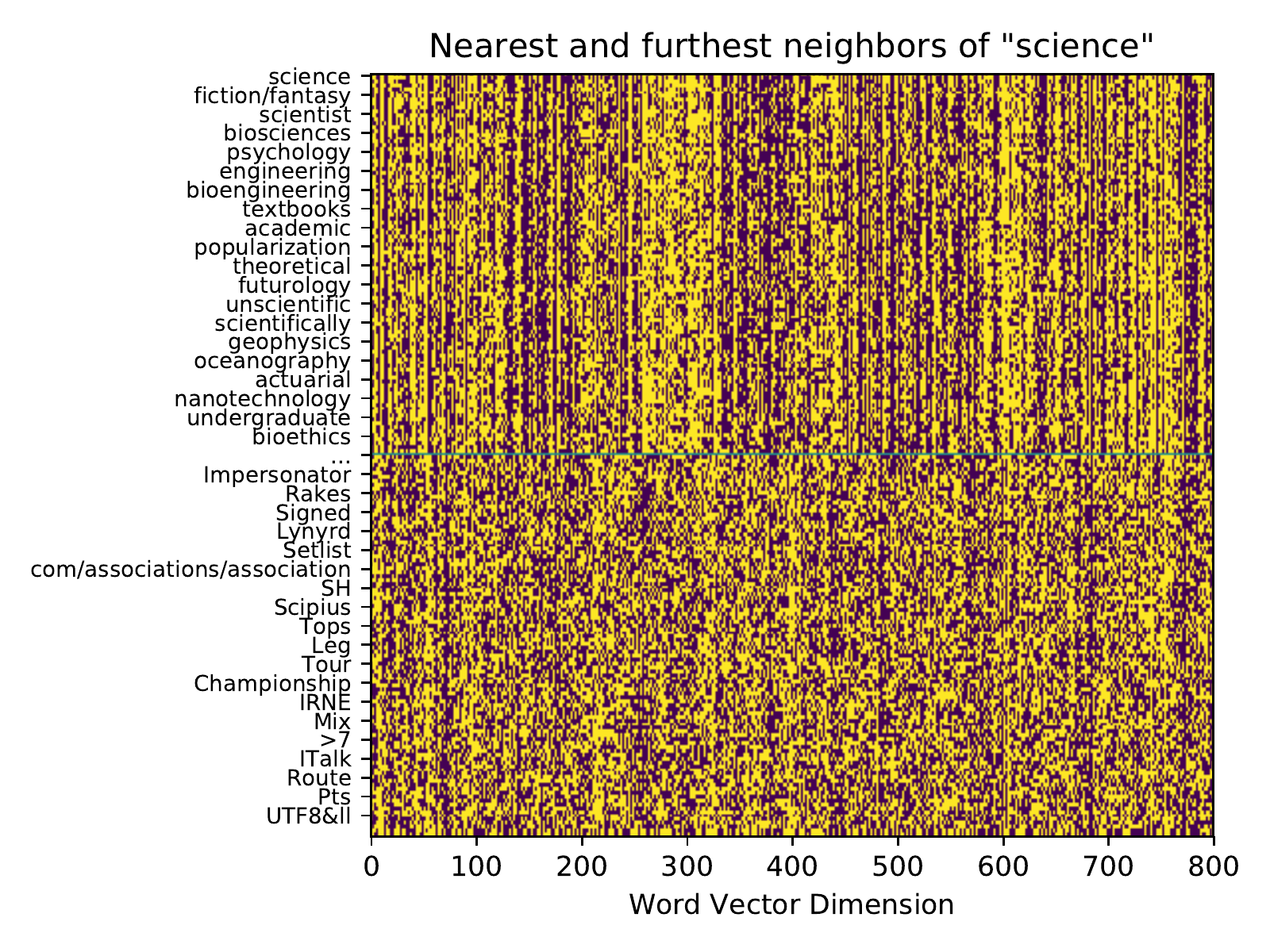}}}%
    \\
    \subfloat[]{{\includegraphics[width=6cm]{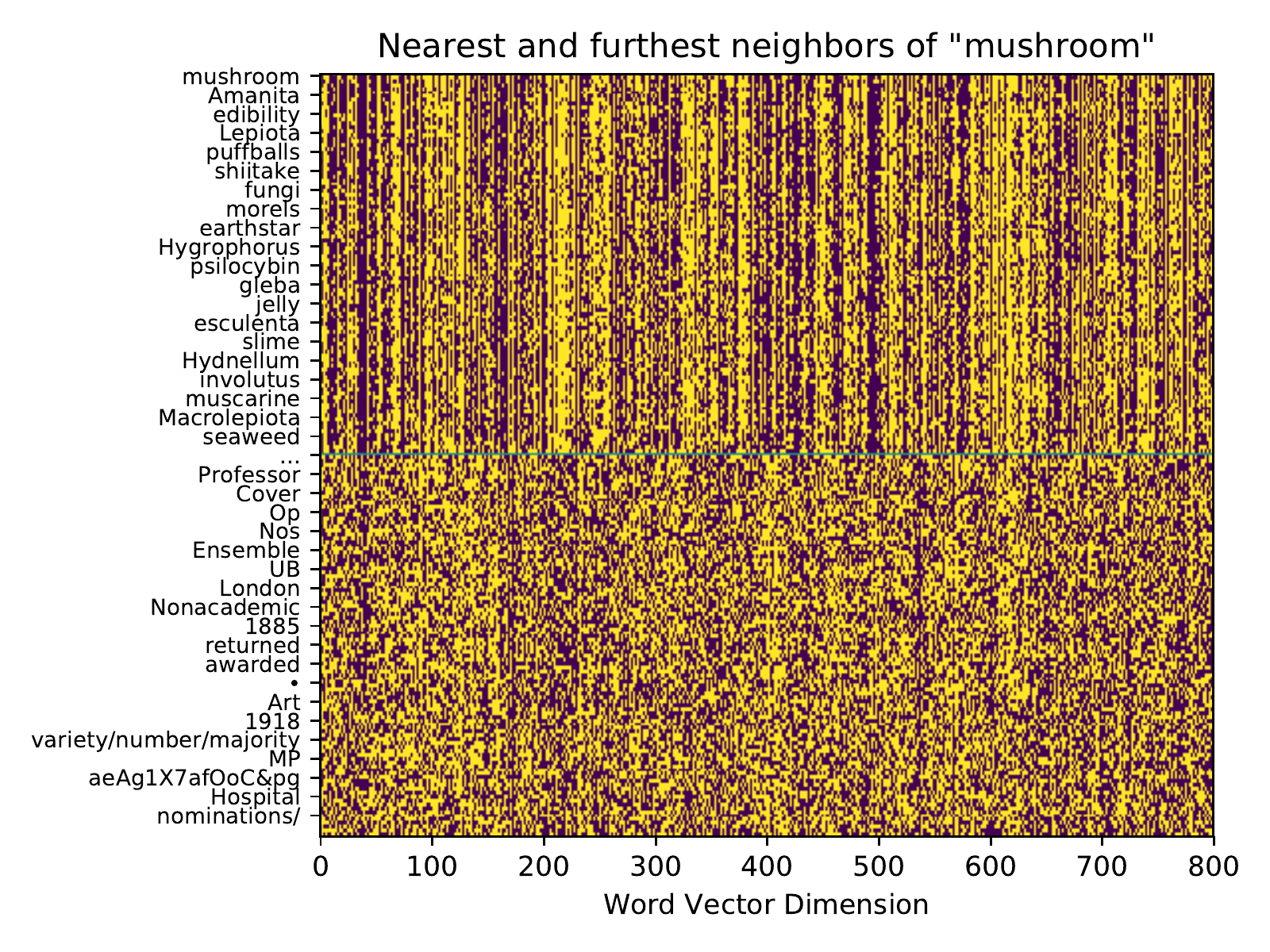}}}%
    \qquad
    \subfloat[]{{\includegraphics[width=6cm]{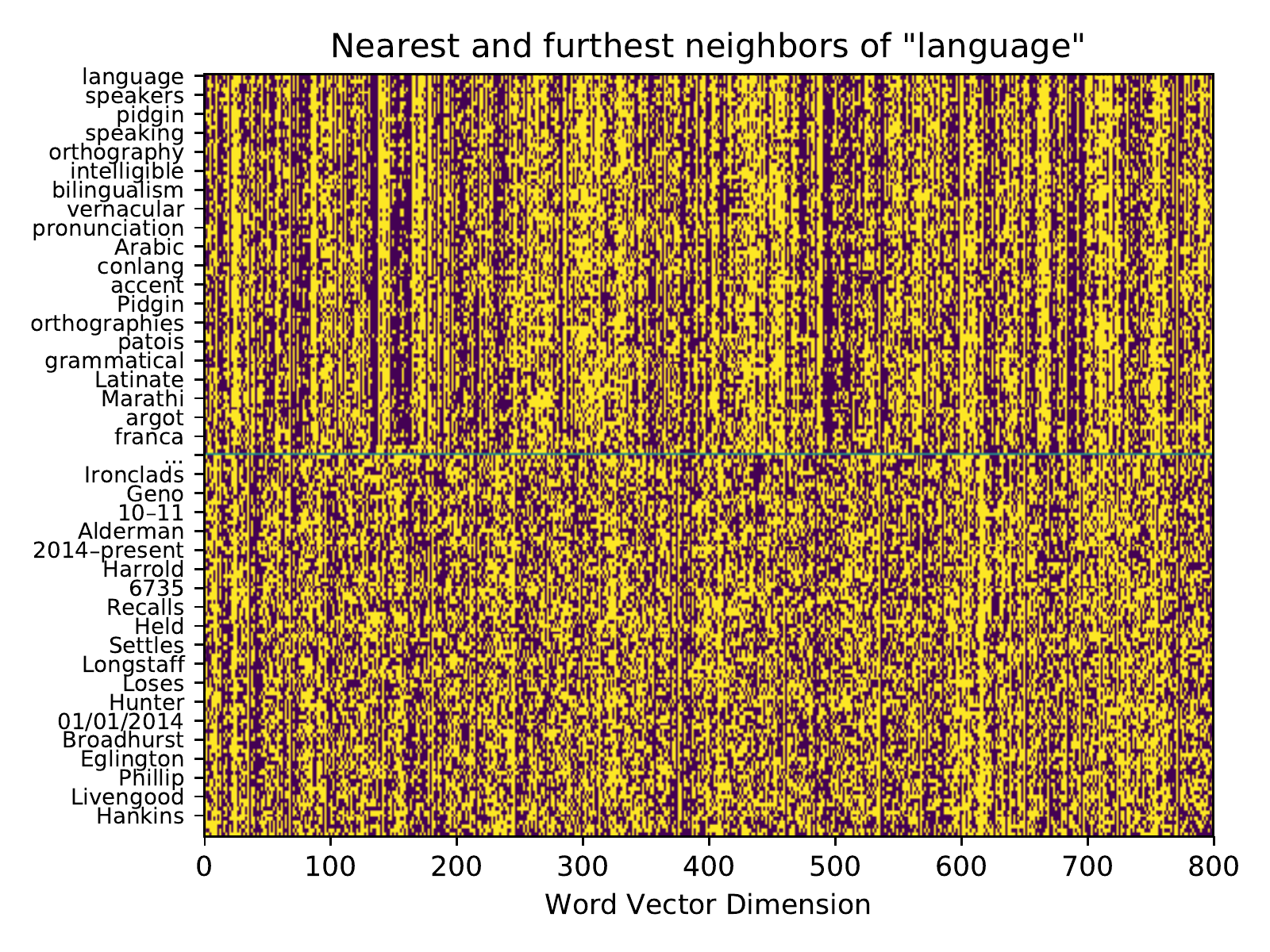}}}%
    \\
    \subfloat[]{{\includegraphics[width=6cm]{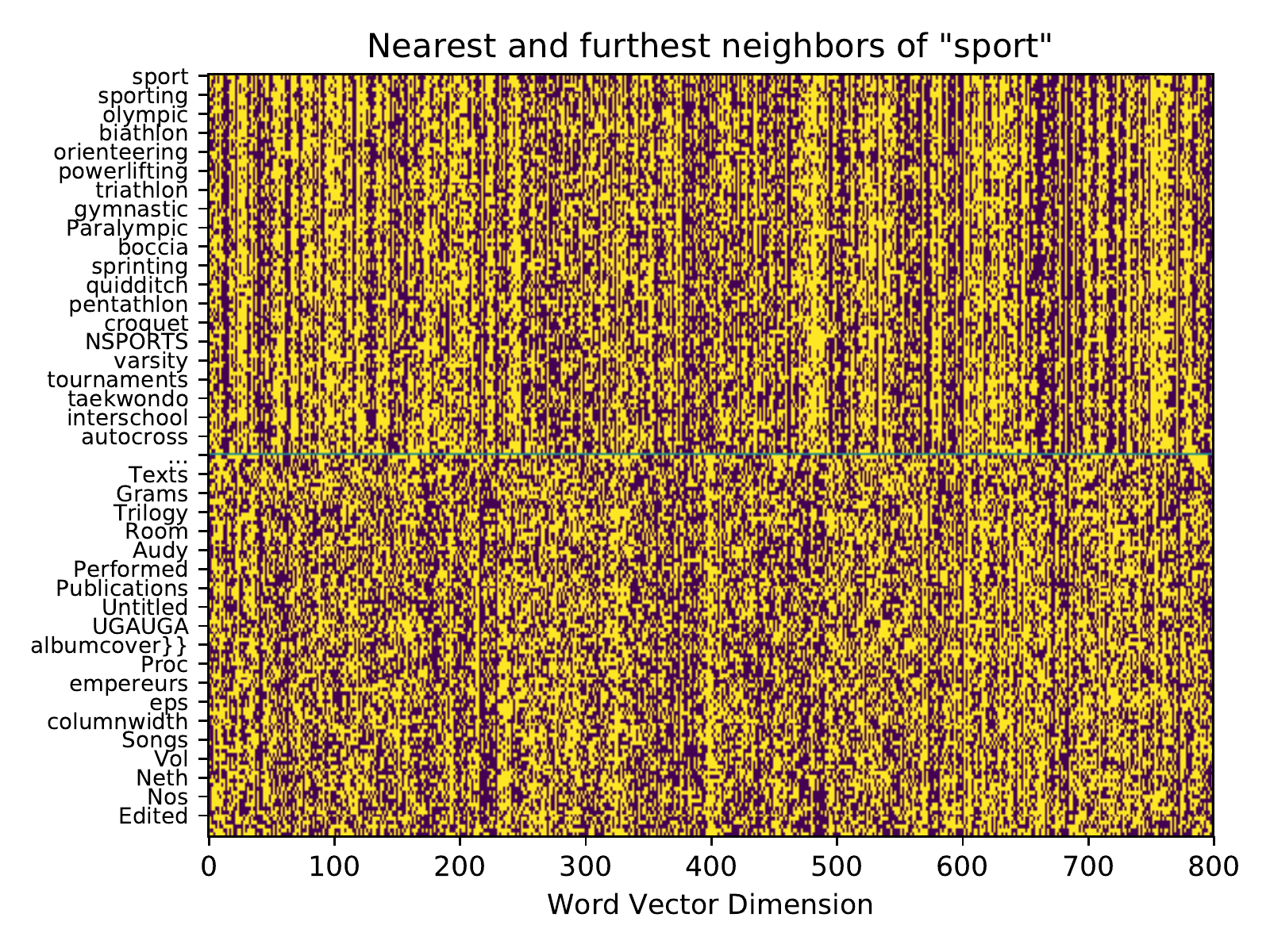}}}%
    \qquad
    \subfloat[]{{\includegraphics[width=6cm]{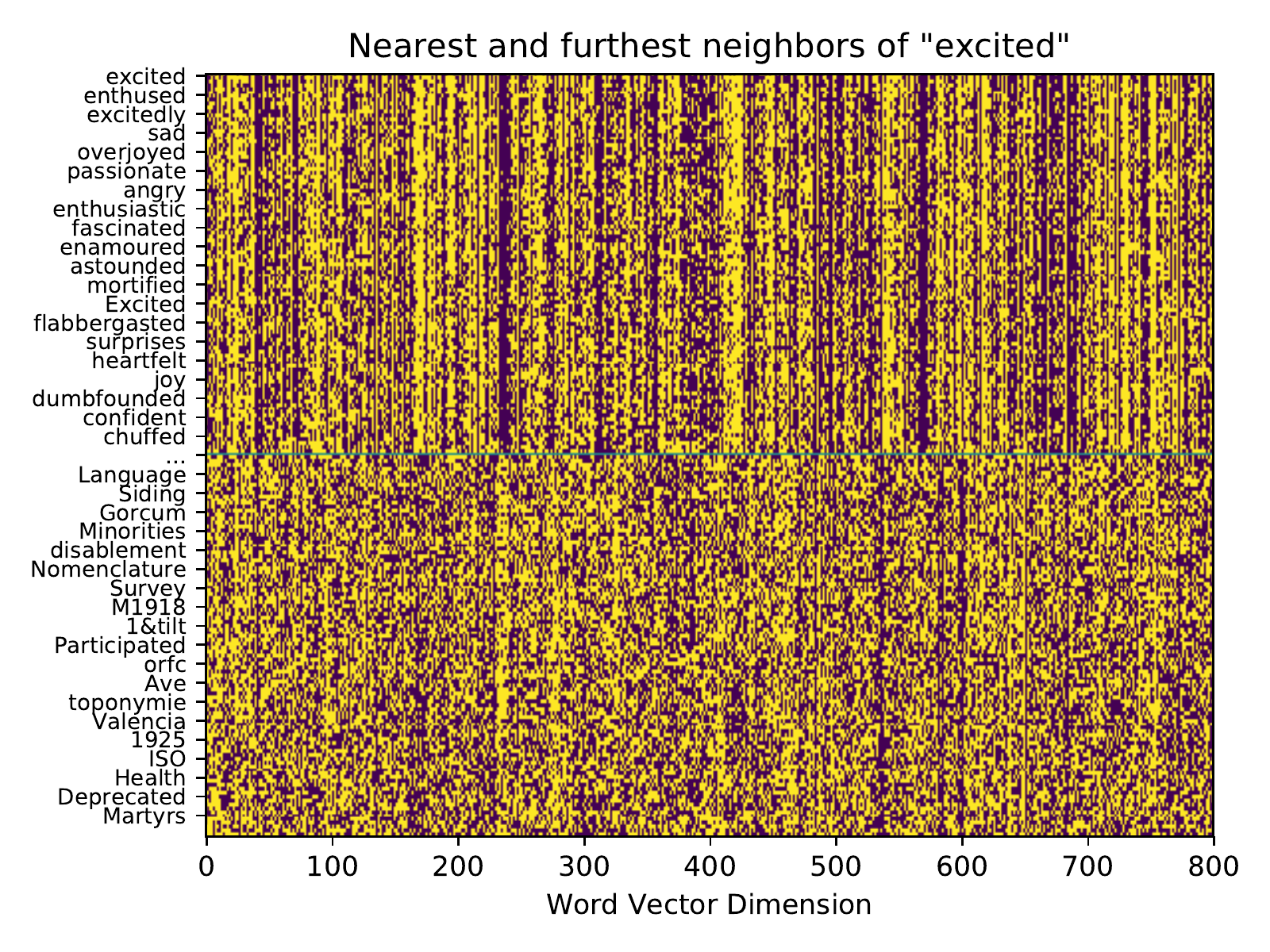}}}%
    \caption{Visualizing 1-bit 800 dimensional word vectors}
    \label{fig:viz}
\end{figure}

\section{Conclusion and Future Work}

In this report we have shown that it is possible to train high quality
quantized vectors that take 8-16x less storage/memory than full
precision vectors. Interestingly, quantized word vectors perform
better than full precision vectors on both word similarity and
question answering, but worse on word analogy. The data suggest that
performing well on word analogy tasks require a higher number of bits
per word while doing well on word similarity tasks require
fewer. Another interesting observation is that performance on the
intrinsic tasks did not really predict performance on extrinsic tasks
(SQuAD) -- this validates the findings of [3, 12]. We  have also shown
that full precision Word2Vec training is prone to overfitting (across
training epochs and across word vector dimension) on smaller datasets
(100MB of Wikipedia); this suggests it is not always better to train
for many epochs. We believe the same phenomena holds for larger
datasets. A final interesting observation is that parameter values of
Word2Vec vectors tend to ``explode'' with higher dimensions, an issue
that virtually quantized training does not have. This suggests it may
be helpful to introduce a regularization term to Word2Vec.

Future work involves evaluating full precision vectors and quantized
vectors on other extrinsic tasks, which will give a more complete
picture of the relative performance of the two. We would also like to
train quantized word vectors on much larger corpuses of data such as
Common Crawl or Google News. Another task is to validate that
overfitting occurs on larger datasets (full English Wikipedia) with
respect to various tasks (other intrinsic tasks, extrinsic
tasks). Finally, we believe it is possible to do virtually quantized
training with Glove, though initial experiments suggest that
several modifications to the loss function are needed to make it work.

\subsubsection*{Acknowledgments}

Huge thanks to Sen Wu, Christopher Aberger, Kevin Clark and Professors
Richard Socher and Christopher R\'e for their advice and assistance on
the project!

\subsubsection*{References}

\small{ [1] Armand Joulin, Edouard Grave, Piotr Bojanowski, Matthijs
  Douze, Herve Jegou, and Tomas Mikolov. 2016. Fast-text.zip:
  Compressing text classification models. {\it arXiv preprint
  arXiv:1612.03651.}}

\small{[2] Armand Joulin, Edouard Grave, Piotr Bojanowski, and
  Tomas Mikolov. 2016. Bag of tricks for efficient text
  classification. {\it arXiv preprint arXiv:1607.01759.}
}

\small{[3] Billy Chiu, Anna Korhonen, and Sampo Pyysalo.
  2016. Intrinsic evaluation of word vectors fails
  to predict extrinsic performance. {\it Proceedings of RepEval 2016.}
}

\small{[4] Omer Levy, Yoav Goldberg, and Ido Dagan. 2015. Improving
  distributional similarity with lessons learned from word
  embeddings. {\it TACL.}
}

\small{[5] Y. LeCun, J. S. Denker, S. Solla, R. E. Howard, and
  L. D. Jackel. Optimal brain damage. In {\it Advances in Neural
    Information Processing Systems}, pages 598-605, 1990.
}

\small{[6] Han, Song, Pool, Jeff, Tran, John, and Dally, William
  J. Learning both weights and connections for efficient neural
  networks. In {\it Advances in Neural Information Processing Systems,
  2015}.
}

\small{[7] Tomas Mikolov, Ilya Sutskever, Kai Chen, Greg Corrado, and
  Jeff Dean. 2013. Distributed representations of words and phrases
  and their compositionality. {\it In Advances in Neural Information
  Processing Systems} 26, pages 3111-3119.}

\small{[8] Tomas Mikolov, Kai Chen, Greg Corrado, and Jeffrey
  Dean. Efficient estimation of word representations in vector
  space. In {\it International Conference on Learning Representations:
    Workshops Track}, 2013.  arxiv.org/abs/1301.3781.}

\small{[9] Courbariaux, M., Hubara, I., Soudry, D., El-Yaniv, R., and
  Bengio, Y. Binarized Neural Networks: Training Deep Neural Networks
  with Weights and Activations Constrained to +1 or -1. {\it arXiv preprint
  arXiv:1602.02830}, 2016.

\small{[10] Hinton, Geoffrey. Neural networks for machine
  learning. Coursera, video lectures, 2012.}

\small{[11] Pennington, Jeffrey, Richard Socher, and Christopher D
  Manning. 2014. Glove: Global vectors for word representation. {\it
    Proceedings of the Empiricial Methods in Natural Language
    Processing (EMNLP 2014)} 12.}

\small{[12] Tobias Schnabel, Igor Labutov, David Mimno, and Thorsten
  Joachims. 2015. Evaluation methods for unsupervised word embeddings.
  In {\it Proc. of EMNLP.}}

\small{[13] P. Rajpurkar, J. Zhang, K. Lopyrev, and P. Liang. Squad:
  100,000+ questions for machine comprehension of text. In {\it Empirical
  Methods in Natural Language Processing (EMNLP)}, 2016.}

\small{[14] Chen, D.; Fisch, A.; Weston, J.; and Bordes, A. 2017.
  Reading wikipedia to answer open-domain questions. {\it arXiv preprint
  arXiv:1704.00051}}

\small{[15] Lev Finkelstein, Evgeniy Gabrilovich, Yossi Matias, Ehud
  Rivlin, Zach Solan, Gadi Wolfman, and Eytan Ruppin. 2002. Placing
  search in context: The concept revisited. {\it ACM Transactions on
    Information Systems}, 20(1):116-131.}

\small{[16] Torsten Zesch, Christof Muller, and Iryna Gurevych.
  2008. Using wiktionary for computing semantic relatedness. In {\it
    Proceedings of the 23rd National Conference on Artificial
    Intelligence - Volume 2}, AAAI'08, pages 861-866. AAAI Press.}

\small{[17] Eneko Agirre, Enrique Alfonseca, Keith Hall, Jana
  Kravalova, Marius Pasca, and Aitor Soroa. 2009.  A study on
  similarity and relatedness using distributional and wordnet-based
  approaches. In {\it Proceedings of Human Language Technologies: The
    2009 Annual Conference of the North American Chapter of the
    Association for Computational Linguistics}, pages 19-27, Boulder,
  Colorado, June. Association for Computational Linguistics.}

\small{[18] Elia Bruni, Gemma Boleda, Marco Baroni, and Nam Khanh
  Tran. 2012. Distributional semantics in technicolor. In {\it
    Proceedings of the 50th Annual Meeting of the Association for
    Computational Linguistics (Volume 1: Long Papers)}, pages 136-145,
  Jeju Island, Korea, July. Association for Computational
  Linguistics.}

\small{[19] Kira Radinsky, Eugene Agichtein, Evgeniy Gabrilovich, and
  Shaul Markovitch. 2011. A word at a time: Computing word relatedness
  using temporal semantic analysis. In {\it Proceedings of the 20th
  international conference on World wide web, pages 337-346. ACM.}}

\small{[20] Minh-Thang Luong, Richard Socher, and Christopher
  D. Manning. 2013. Better word representations with recursive neural
  networks for morphology.  In {\it Proceedings of the Seventeenth
  Conference on Computational Natural Language Learning}, pages
  104-113, Sofia, Bulgaria, August. Association for Computational
  Linguistics.}

\small{[21] Roi Reichart Felix Hill and Anna Korhonen. 2014.
  Simlex-999: Evaluating semantic models with (genuine) similarity
  estimation. {\it arXiv preprint arXiv:1408.3456.}}

\small{[22] Courbariaux, M., Bengio, Y., David, J.P.: Training deep
  neural networks with low precision multiplications. {\it arXiv preprint
  arXiv:1412.7024 (2014)}}

\small{[23] W. Chen, J. T. Wilson, S. Tyree, K. Q. Weinberger, and
  Y. Chen, ``Compressing neural networks with the hashing trick,'' in
  {\it International Conference on Machine Learning (ICML)}, 2015,
  pp. 2285-2294.}

\small{[24] Abigail See, Minh-Thang Luong, and Christopher
  D. Manning. 2016. Compression of Neural Machine Translation via
  Pruning. In {\it Proceedings of CoNLL.}}

\small{[25] Raphael Shu, Hideki Nakayama. 2017. Compressing Word
  Embeddings via Deep Compositional Code Learning. In {\it
    International Conference on Learning Representations}, 2017.}

\small{[26] Christopher De Sa, Megan Leszczynski, Jian Zhang, Alana
  Marzoev, Christopher R. Aberger, Kunle Olukotun, Christopher
  R\'e. 2018. High-Accuracy Low-Precision Training. {\it arXiv
    preprint arXiv:1803.03383}}

\small{[27] Bryan McCann, James Bradbury, Caiming Xiong, and Richard
  Socher. Learned in translation: Contextualized word vectors. In {\it
    NIPS}, 2017.}

\small{[28] Neelakantan, Arvind, Vilnis, Luke, Le, Quoc V., Sutskever,
  Ilya, Kaiser, Lukasz, Kurach, Karol, and Martens, James. Adding
  gradient noise improves learning for very deep networks. {\it ICLR
  Workshop}, 2016.}

\small{[29] Hinton, G. Vinyals, O. and Dean, J. Distilling knowledge
  in a neural network. In {\it Deep Learning and Representation
    Learning Workshop, NIPS}, 2014.}

\small{[30] Bengio, Yoshua. Estimating or propagating gradients
  through stochastic neurons. {\it Technical Report arXiv:1305.2982,
  Universite de Montreal}, 2013.}

\small{[31] Matt Mahoney. About the test data, 2011. URL
  http://mattmahoney.net/dc/textdata.}

\small{[32] B. Hassibi, D. Stork, G. Wolff and T. Watanabe. Optimal
  brain surgeon: Extensions and performance comparisons. In {\it Advances
    in Neural Information Processing Systems 6}.  Morgan Kaufman, San
  Mateo, CA: 263-270, 1994.
}

\end{document}